\def\year{2021}\relax
\PassOptionsToPackage{table}{xcolor}
\documentclass[letterpaper, table]{article} 
\pdfoutput=1
\usepackage{aaai21}  
\usepackage{times}  
\usepackage{helvet} 
\usepackage{courier}  
\usepackage[hyphens]{url}  
\usepackage{graphicx} 
\usepackage{natbib}  
\usepackage{caption} 
\usepackage{multirow}
\usepackage{tabularx}
\usepackage{listings}
\usepackage{booktabs}
\usepackage[switch]{lineno}
\usepackage[dvipsnames]{xcolor}

\title{Automated Lay Language Summarization of Biomedical Scientific Reviews}

\author {
        Yue Guo,\thanks{These authors contributed equally to this work.}\textsuperscript{\rm 1}
        Wei Qiu, \footnotemark[1]\textsuperscript{\rm 2} Yizhong Wang, \textsuperscript{\rm 2}
        Trevor Cohen\textsuperscript{\rm 1}\\
}
\affiliations {
    \textsuperscript{\rm 1} Biomedical and Health Informatics, University of Washington  \\
    \textsuperscript{\rm 2} Paul G. Allen School of Computer Science, University of Washington\\
    yguo50@uw.edu, \{wqiu0528, yizhongw\}@cs.washington.edu, cohenta@uw.edu 
}

\begin{document}
\newcolumntype{L}[1]{>{\raggedright\arraybackslash}p{#1}}
\newcolumntype{C}[1]{>{\centering\arraybackslash}p{#1}}
\newcolumntype{R}[1]{>{\raggedleft\arraybackslash}p{#1}}
\maketitle

\begin{abstract}
Health literacy has emerged as a crucial factor in making appropriate health decisions and ensuring treatment outcomes. However, medical jargon and the complex structure of professional language in this domain make health information especially hard to interpret. Thus, there is an urgent unmet need for automated methods to enhance the accessibility of the biomedical literature to the general population. This problem can be framed as a type of translation problem between the language of healthcare professionals, and that of the general public.  In this paper, 
we introduce the novel task of automated generation of lay language summaries of biomedical scientific reviews, and construct a dataset to support the development and evaluation of automated methods through which to enhance the accessibility of the biomedical literature.  We conduct analyses of the various challenges in performing this task, including not only summarization of the key points but also explanation of background knowledge and simplification of professional language. We experiment with state-of-the-art summarization models as well as several data augmentation techniques, and evaluate their performance using both automated metrics and human assessment. Results indicate that automatically generated summaries produced using contemporary neural architectures can achieve promising quality and readability as compared with reference summaries developed for the lay public by experts (best ROUGE-L of 50.24 and Flesch-Kincaid readability score of 13.30). We also discuss the limitations of the current effort, providing insights and directions for future work.
\end{abstract}

\section{Introduction}
The ability to understand scientific concepts, content, and research in the medical domain is defined as health literacy \cite{parker1999health}, which is crucial to making appropriate health decisions and ensuring treatment outcomes. The development of the internet has enabled the general population to access health information, greatly expanding the volume of available health education materials. However, challenges arise in reading and understanding these materials because of the inability to identify credible resources \cite{howes2004evidence}, unfamiliarity with medical jargon \cite{korsch1968gaps}, and the complex structure of professional language \cite{friedman2002two}. Furthermore, knowledge in the health domain evolves over time, presenting laypeople with the additional challenge of discerning the most up-to-date information. The COVID-19 pandemic has cast a spotlight on the challenges in the general public's ability to obtain, interpret, and apply knowledge to guide their health-related behavior. These challenges are exemplified by difficulties in interpreting articles from the biomedical literature, for those without specific training in this domain. Our project aims to bridge the gap between the increasing availability of health information and the difficulty the public has understanding it. To do so, we confront the task of rendering the biomedical literature comprehensible to the lay public, which can be framed as a type of translation problem: from the language of healthcare professionals to plain language.

A systematic review, such as those in the widely-used Cochrane Database of Systematic Reviews\footnote{http:///www.cochranelibrary.com} (CDSR), is a type of biomedical scientific publication that synthesizes current empirical evidence for a research question, to identify the strongest evidence to inform decision making while minimizing bias. Of importance for the current research, reviews in the CDSR include lay language summaries, written by review authors or Cochrane staff. In this paper, we introduce the novel task of automated generation of lay language summaries of biomedical scientific reviews. We introduce a dataset constructed by extracting 7,805 high-quality abstract pairs that consist of both abstracts intended for professional readers, and plain language versions written by domain experts. The \textit{source} in the training dataset is the healthcare professional version of an abstract, with an average length of 714 words. The \textit{target} is the corresponding plain language version, with an average length of 371 words. 

This dataset is the first corpus developed to facilitate development of methods to improve document-level understandability of biomedical narratives for the general public, and covering a broad range healthcare topics. The closest parallel in the literature may be the manually annotated dataset, MSD, which was developed recently to improve bi-directional communication between clinicians and patients \cite{cao2020expertise}. MSD is focused on communication of clinical topics at the sentence level, and the accompanying work approached the problem using text style transfer and simplification algorithms. However, constraining the task to sentence level prohibits methods from considering the broader context in which a sentence occurs. 
While the text summarization community has developed various corpora 
\cite{allahyari2017text} for document-level summarization tasks, current resources for the biomedical scientific domain are limited \cite{Moradi2019TextSI}. Furthermore, the proposed plain language summarization task imposes additional challenges, such as terminology explanation and sentence structure simplification, that are not required for general domain summarization (see Table \ref{transformation_case_study} for an analysis of five additional task components).
One important goal of our work is to meet the need for a dataset to support research into the task of generating summaries of biomedical  professional literature that are comprehensible to the lay public.

To approach this task, we implemented several state-of-the art extractive and abstractive summarization models and evaluated them on the collected CDSR dataset. On account of the limited size of this dataset, we also applied intermediate pre-training (both out-of-domain on-task, and in-domain off-task) to the best abstractive model. We evaluated the utility of pre-training this model on CNN/DM \cite{nallapati2016abstractive}, a much larger general domain summarization dataset. To provide the model with more domain-specific biomedical language, we pre-trained it on an unlabeled biomedical corpus of 300K abstracts from the PubMed database. 

Standard automated metrics of summarization and readability were adopted to evaluate model performance. In addition, we used ratings of human evaluators to assess the generated summaries from several perspectives. The results suggest that the best-performing model can generate lay language summaries with promising quality and readability.

Our main contributions can be summarized as follows:
\begin{itemize}
  \item We introduce the novel task of automated generation of lay language summaries of biomedical scientific reviews.
  \item We construct a dataset of 7,805 summaries and a qualitative analysis of the NLP challenges inherent in this task. 
  \item We evaluate performance of state-of-the-art summarization models leveraging neural machine translation architectures, with and without data augmentation techniques, on this task.
  \item We conduct automated and human evaluation from multiple perspectives, showing that machine generated summaries can achieve promising quality and readability as compared with reference summaries developed for the lay public by domain experts.\footnote{We release our code at \url{https://github.com/qiuweipku/Plain_language_summarization}}
\end{itemize}

\begin{table*}[t!]
\centering
\tiny
\begin{tabular}{@{}m{1.4cm}<{\centering}m{3.55cm}<{\centering}m{3.55cm}<{\centering}m{3.55cm}<{\centering}m{3.55cm}<{\centering} @{}}
\toprule
\textbf{Category}  & \textbf{Source}   & \textbf{Target}  & \textbf{BART+CNN/DM+PubMed} & \textbf{BART+PubMed}  \\ \midrule
Removing Unnecessary Details 
& \ldots \textcolor{magenta}{A complete case analysis (i.e. participants who completed the study) among trials investigating CDAD (31 trials, 8672 participants)} suggests that probiotics reduce the risk of CDAD by 60\%. \textcolor{magenta}{The incidence of CDAD was 1.5\% (70/4525) in the probiotic group compared to 4.0\% (164/4147) in the placebo or no treatment control group (RR 0.40, 95\% CI 0.30 to 0.52; GRADE = moderate)}. \ldots \cite{goldenberg2017probiotics} 
& \ldots Our results suggest that when probiotics are given with antibiotics the risk of developing CDAD is reduced by 60\% on average. \ldots
& \ldots We found that probiotics reduce the risk of developing CDAD by 60\% \textcolor{magenta}{(1.5\% (70/4525) versus 4.0\% (164/4147) in the placebo or no treatment control group; GRADE = moderate)}. \ldots
& \ldots \textcolor{magenta}{A complete case analysis (i.e. participants who completed the study) among trials investigating CDAD (31 trials, 8672 participants)} suggests that probiotics reduce the risk of developing CDAD by 60\%. \textcolor{magenta}{A total of 1.5\% (70/4525) of participants in the probiotic group developed CDAD compared to 4.0\% (164/4147) in the placebo or no treatment control group (moderate quality evidence)}. \ldots
\\ \midrule
Relevant Background Explanation 
& Hepatitis C is a major cause of liver‐related morbidity and mortality. Standard therapy is ribavirin plus pegylated interferon to achieve undetectable level of virus in the blood, but the effect on clinical outcomes is controversial. \ldots \cite{brok2010ribavirin}
& \textcolor{teal}{Globally about 170 million people are chronically infected with hepatitis C virus. Hepatitis C is a blood‐borne virus and routes of transmission include intravenous drug use, mother‐to‐infant transmission, unsafe medical practices, high‐risk sexual behavior, and blood transfusion. Chronic hepatitis C is in most patients a benign viral infection, but a minority of patients develop liver cirrhosis and may suffer from complications due to cirrhosis or die. It is known that treatment with the drug interferon clears hepatitis C virus from the blood in about 15\% of patients}. \ldots
& Hepatitis C is a major cause of liver‐related morbidity and mortality. Standard therapy is ribavirin plus pegylated interferons to achieve undetectable level of virus in the blood, but the effect on clinical outcomes is controversial. \ldots
& Hepatitis C is a major cause of liver‐related morbidity and mortality. Standard therapy is ribavirin plus pegylated interferons to achieve undetectable level of virus in the blood, but the effect on clinical outcomes is controversial. \ldots
\\ \midrule
Jargon Explanation
& \ldots We considered all \textcolor{violet}{randomised controlled trials (RCTs)} comparing EVLA, endovenous RFA or UGFS with conventional surgery in the treatment of SSV varices for inclusion. \ldots \cite{paravastu2016endovenous} 
& \ldots We found three \textcolor{violet}{randomised controlled trials (clinical studies where people are randomly put into one of two or more treatment groups)} that compared endovenous lasers (EVLA) with surgery.  \ldots
& \ldots We found three \textcolor{violet}{randomised controlled trials (clinical studies where people are randomly put into one of two or more treatment groups)} that compared endovenous lasers (EVLA) with surgery. \ldots
& \ldots We found three \textcolor{violet}{randomised controlled trials (clinical studies where people are randomly put into one of two or more treatment groups)} that compared endvenous laser Ablation with surgery. \ldots
 \\ \midrule
Converting to Interrogative Sentence 
& \ldots Non-resection versus resection for an asymptomatic primary tumour in patients with unresectable stage IV colorectal cancer \ldots \cite{cirocchi2012non}  
& \ldots Should the primary cancer be surgically removed in asymptomatic patients with unresectable stage IV colorectal cancer? \ldots
& \ldots Primary tumour resection versus no resection in asymptomatic patients with unresectable stage IV colorectal cancer who are treated with palliative chemo/radiotherapy 
& Primary tumour resection in asymptomatic patients with unresectable stage IV colorectal cancer who are treated with palliative chemo/radiotherapy \ldots
\\ \midrule
Sentence Structure Simplification 
& \ldots Abnormal blood flow patterns in fetal circulation \textcolor{orange}{detected by} Doppler ultrasound may \textcolor{cyan}{indicate poor fetal prognosis}. \ldots \cite{alfirevic2017fetal} 
& \ldots Doppler ultrasound \textcolor{orange}{detects} changes in the pattern of blood flow through the baby's circulation. \textcolor{cyan}{These changes may identify babies who have problems.} \ldots
& \ldots Abnormal blood flow patterns in the blood vessels of the unborn baby may \textcolor{cyan}{indicate poor outcomes for the baby.}  \ldots  
& \ldots The aim of the review was to find out if using Doppler ultrasonography ( ultrasound of the baby's heart and blood vessels) during pregnancy can improve outcomes for babies. \ldots \\ 
\bottomrule
\end{tabular}
\caption{Typical transformation phenomena from \textit{source} to \textit{target}, and the corresponding generated summaries using two best-performing models. We only show part of the long text here for brevity.}
\label{transformation_case_study}

\end{table*}

\section{Related Work}
\subsection{Scientific Document Summarization}
Automated summarization of scientific documents has been a long-standing research topic \cite{Paice1980TheAG,teufel2002summarization}. Advances have been achieved through the development of high-quality datasets and evaluation tasks, including but not limited to abstract generation of academic papers \cite{Cohan2018ADA}, citation sentence generation \cite{Luu2020CitationTG} or extreme
summarization of the entire document \cite{Cachola2020TLDRES}. In the medical field, \citet{Sarkar2011UsingML} explored extractive summarization of medical news articles. 
The Text Analysis Conference (TAC) 2014 Biomedical Summarization track introduced several subtasks to evaluate citation-based summaries. 
Our proposed task differs from this prior work in two ways: 1) our target summaries are in plain language, so the overall task requires other capabilities beyond summarization; 
2) the target summaries in our dataset are considerably longer (see Table \ref{data_description}), which poses further challenges.

Existing summarization methods can be broadly categorized into extractive and abstractive approaches \cite{Andr2007ASO}. The former \cite{Erkan2004LexRankGC, Cheng2016NeuralSB} select sentences from the original text, while the latter \cite{rush2015abstractive, nallapati2016abstractive} can generate summaries using words that are not found in the original document. Our task is abstractive by nature. For scientific document summarization, various features (e.g. citation networks) can be used to improve performance. While augmentation of this sort is beyond the scope of the current work, we refer the interested reader to \citet{Altmami2020AutomaticSO} and \citet{ Moradi2019TextSI} for a  comprehensive overview of the relevant approaches.
\subsection{Text Simplification} 
Text simplification \cite{Shardlow2014ASO} modifies the content or structure of a text to make it easier to understand. Unlike summarization, text simplification approximates the meaning of the original sentence without necessarily shortening it.
Simplification techniques have been used in the biomedical domain for generating patient-centered radiology reports \cite{qenam2017text}, name entity recognition \cite{habibi2017deep}, preprocessing for biomedical interaction recognition \cite{baumgartner2008concept}, syntactic parsing \cite{jonnalagadda2010towards}, and simplification of medical journal text \cite{jonnalagadda2010towards}. For lexical simplification, WordNet \cite{miller1995wordnet}, the UMLS \cite{bodenreider2004unified} and Wiktionary \cite{zesch2008extracting} are widely used as synonym resources to find and replace medical terms. Toward the goal of producing summaries of abstracts that are understandable to a lay audience, our task requires simplification at the document rather than the single sentence level, and combines this with summarization to achieve both shorter and more readily 
understandable summaries.
\subsection{Pre-training and Transfer Learning}
Large pre-trained neural networks have led to recent advances in performance across a broad range of NLP tasks \cite{peters2018elmo, Radford2018ImprovingLU,Devlin2019BERTPO} - especially when the available labeled data are limited \cite{brown2020language}, or there is a shift in domains involved \cite{Hendrycks2020PretrainedTI}. Due to the complexity of our task and the relatively small size of our data, we posit that pre-training is a prerequisite to strong performance. Furthermore, recent work shows that adaptive pre-training with domain-relevant unlabeled data \cite{Gururangan2020DontSP} or task-relevant labeled data \cite{Pruksachatkun2020IntermediateTaskTL} can further improve the performance of pre-trained models. The intermediate pre-training strategies we evaluated in our experiments are inspired by these findings, and we encourage efforts in finding additional useful data to further improve performance on this task.
\subsection{Biomedical Domain Summarization}
The most common document types used for summarization tasks in the biomedical domain are clinical notes, with the aim to reduce information overload for health practitioners \cite{pivovarov2015automated, feblowitz2011summarization, molla2011development}. This aim differs from our main objective: generating lay language summaries of biomedical scientific reviews for health consumers. Another related area of work concerns information retrieval from the internet, where the goal is to help consumers find (rather than interpret) health information \cite{goeuriot2020overview}. 




\section{Dataset Description}
\subsection{The CDSR}
The Cochrane Database of Systematic Reviews (CDSR) includes high-quality systematic reviews \cite{uman2011systematic} in various health care domains that facilitate evidence-based medical decision making. For a systematic review, two independent reviewers will review eligible peer-reviewed papers, registered clinical trials, conference papers, or ``grey literature''\footnote{This can be loosely defined as literature that is disseminated outside the usual publishing channels.}; search for evidence on a clearly formulated question; extract data from the studies; and grade the quality of available data. According to the hierarchy of scientific evidence \cite{murad2016new}, systematic review is the most robust evidence supporting an argument. Of particular importance for the current work, a plain language version of each abstract accompanies its professional language counterpart. Of note, plain language summaries have been required from authors submitting a review since 2015. Prior to this, they were written by Cochrane staff with specialized training.
\subsection{Dataset construction}
We extracted 7,805 abstracts (\textit{source}), paired with their plain language versions (\textit{target}) from CDSR reviews available up to March 29, 2020. The original data is downloadable via the official API \footnote{\url{https://www.cochranelibrary.com/cdsr/reviews}}. We only retained examples with source length between 300 to 1,000 words, and target length between 100 to 700 words. This resulted in a set of 5,195 source-target pairs which constitutes our training set, a further 500 abstract pairs as the validation set, and 1000 more as the test set. 
\subsection{Data analysis}
\label{Data_analysis}

\begin{table*}
\centering
\begin{tabular}{@{} l rr rr rr @{}}
\toprule
& \multicolumn{2}{c}{\bf Train} & \multicolumn{2}{c}{\bf Validation} & \multicolumn{2}{c}{\bf Test} \\ 
\cmidrule(lr){2-3}
\cmidrule(lr){4-5}
\cmidrule(lr){6-7}
\multicolumn{1}{c}{}                           & \textbf{Source}       & \textbf{Target}      
& \textbf{Source}       & \textbf{Target}          
& \textbf{Source}       & \textbf{Target}      \\ 
\midrule
\# abstracts                              & 5,195        & 5,195       & 500            & 500            & 1,000       & 1,000       \\
Average length (words)                      & 714          & 374         & 713            & 368            & 727         & 378         \\
Vocabulary size                                  & 57,685       & 34,175       & 16,574         & 10,596         & 23,938      & 15,407      \\ \midrule
Flesch-Kincaid & 14.68 & 13.25 & 14.93 & 13.57 & 14.70 & 13.23\\
Gunning & 14.57 & 13.54 & 14.69 & 13.79 & 14.57 & 13.49\\
Coleman-Liau & 15.40 & 14.37 & 15.57 & 14.51 & 15.39 & 14.43 \\
\bottomrule

\end{tabular}
\caption{Dataset statistics across the different splits.}
\label{data_description}

\end{table*}


Table \ref{data_description} shows the characteristics of the dataset. Methods of calculating the readability score are detailed in the Evaluation Metrics Section\ref{evaluation_metrics}. The readability of both source and target texts are at undergraduate level (e.g. 13th grade = 1st year, 15th grade = 3rd year). Notably, the average lengths of abstracts from the source sets are larger than those from the target sets, and the target sets have lower readability scores (i.e. more readable) on average. Since the length of the abstracts and the readability scores from different subset splits are similar, the dataset splits can be considered to be comparable. 



In order to understand how experts translate scientific biomedical abstracts into plain language versions that target the general population, we identified five typical transformation phenomena on the basis of our observations when constructing the current dataset.\footnote{More comprehensive guidelines for writing Cochrane plain language summaries can be found in \citet{mcilwain2014standards}.} These transformation categories with examples are presented in Table \ref{transformation_case_study}. 

The most common transformation to make the paragraph more straightforward is to remove unnecessary details. Although some details such as experimental settings, control experiments or quantitative results are informative for professionals and may indicate the quality of a scientific review, such information
may confuse laypeople and obscure the key findings from the review. 
The critical message for laypeople is the general association between an intervention and a health condition, rather than the precise details of the scientific evidence used to support this conclusion. 

Explaining relevant background information, including the prevalence, mortality, risk factors and outcome of a condition, enables readers to establish whether or not the topic under discussion meets their information needs. Jargon (or even some standard medical terms) presents another challenge that prevents laypeople from referring to peer-reviewed papers for answers to their health-related questions. Providing definitions for technical terms (such as ``randomized control trials'' in Table \ref{transformation_case_study}) can make professionally-authored text more understandable to a lay audience. Restating the sentence, especially title or headings, in an interrogative sentence makes the scientific content more engaging, and highlights the clinical question under consideration in the review. Sometimes, it is difficult for laypeople to understand the importance of the study question. 

Finally, many cases in our dataset require sentence structure simplification. Rephrasing lengthy, convoluted sentences as shorter ones can divide complex information into smaller, easier-to-process units. We also identified other less frequent transformations, such as avoiding passive voice, using ``must'' to indicate requirements, and minimizing abbreviations. As it would be intractable to exhaustively identify and categorize all of the transformation types in our dataset, we provide only the most commonly encountered ones in this paper.


\section{Evaluation Metrics}
\label{evaluation_metrics}

The various phenomena in our task present a challenge for comprehensive and fair evaluation. Therefore, we adopt several automatic evaluation metrics, as well as human evaluation to assess different aspects of model performance. 
\subsection{Automated evaluation}
\subsubsection{Summarization evaluation}
We first use ROUGE \cite{lin2004rouge} to evaluate the summarization performance. 
ROUGE-n measures overlap of n-grams between the model-generated summary and the human-generated reference summary, and ROUGE-L measures the longest matching sequence of words using the longest common subsequence. 
In this task, we report the F1 scores of ROUGE-1, ROUGE-2, and ROUGE-L as the summarization performance measures. ROUGE scores were computed using \texttt{pyrouge}\footnote{https://pypi.org/project/pyrouge/}.
\subsubsection{Readability evaluation} Other than how much information is retained in the summary, we are also interested in assessing the ease with which a reader can understand a passage, defined as readability. We use three standard metrics to evaluate readability: Flesch-Kincaid grade level \cite{kincaid1975derivation}, Gunning fog index \cite{gunning1952technique}, and Coleman-Liau index \cite{coleman1975computer}.
These scores are computed using \texttt{textstat}\footnote{https://pypi.org/project/textstat/}, and their formulae are as follows:

\begin{itemize}
    \item \textbf{Flesch-Kincaid grade level}:
    $$0.39 \left (\frac{\mbox{total words}}{\mbox{total sentences}}\right ) + 11.8 \left (\frac{\mbox{total syllables}}{\mbox{total words}} \right ) - 15.59,$$
    \item \textbf{Gunning fog index}:
    $$0.4\left[\left(\frac{\mbox{words}}{\mbox{sentences}}\right ) +100\left (\frac{\mbox{complex words}}{\mbox{words}}\right)\right],$$
    where complex words are those words with three or more syllables. 
    \item \textbf{Coleman-Liau index}:
    $$0.0588L - 0.296S - 15.8,$$
    where $L$ is the average number of letters per 100 words and $S$ is the average number of sentences per 100 words. 
\end{itemize} 
    These readability evaluation metrics all estimate the years of education generally required to understand the text. Lower scores indicate that the text is easier to read.
    More specifically, scores of 13-16 correspond to college-level reading ability in the United States education system. Table \ref{data_description} shows the readability scores of the source and target sets of our dataset. Although all the scores indicate a college level of education is needed to read even the target summary, we do see a stable difference between the scores of the source and target. This indicates that these scores are useful for reflecting the different level of readability for text in our dataset.
\subsection{Human evaluation}
While we have adopted the most commonly used metrics for assessing summarization and simplification performance, many aspects of the generated text, such as fluency, grammaticality, and coherence, are not captured by them. Of particular importance, factual correctness of the generated text is crucial in the medical domain. To consider these desirable properties, we developed a method for further assessment of summary quality by human evaluators.

Specifically, we presented an evaluator with two biomedical abstracts followed by four questions. Evaluators were recruited if they: (1) were able to read and write in English; and (2) had at least 12 years education (as the education level required for the training dataset we preprocessed is college level). Evaluators were excluded if they (1) had participated in medical training or shadowing in a hospital; or (2) had completed advanced (graduate level) biology courses. These criteria were selected to ensure that our evaluators were representative of the college-educated lay public. Participants were recruited by convenience sampling, and the study was considered exempt upon institutional IRB review. The estimated time for completion of the human evaluation was 30 minutes for each participant. No compensation was provided for participating in this study.  We recruited 8 human raters. 
The average age of these evaluators was 23.5 years old. Four of them were female, and all had more than 12 years of formal education. Each of the abstract/summary pairs was assigned to two independent evaluators.

Two versions of each biomedical abstract were presented: SOURCE refers to the original professional language version, and SUMMARY refers to the version to be evaluated. This was either the professionally written target, or the version generated by our best-performing automated summarization model, BART pre-trained on both CNN/DM and PubMed. Evaluators were blinded to the authorship of the summary (BART vs. human expert). Two biomedical abstracts (A and B) were randomly selected from the test set. Evaluators were required to read through the two pairs of abstracts and compare the SUMMARY to SOURCE considering the following aspects on a 1-5 Likert scale (1 - very poor; 5 - very good):
\begin{itemize}
    \item \textbf{Grammaticality} Do you think the SUMMARY is grammatically correct? 
    \item \textbf{Meaning preservation} Does the SUMMARY provide all the useful information you think is important from the source? 
    \item \textbf{Understandability} Is the SUMMARY easier to understand than the source?
    \item \textbf{Correctness of key information} How do you judge the overall quality of the SUMMARY in terms of its correctness of the key information compared to the source? 
\end{itemize}


\section{Experiments}
\begin{table*}[thb]
\centering
\begin{tabular}{@{} l ccc ccc @{}}
\toprule
\textbf{Model}                  & \textbf{ROUGE-1}         & \textbf{ROUGE-2}         & \textbf{ROUGE-L}    & \textbf{Flesch-Kincaid} & \textbf{Gunning} & \textbf{Coleman-Liau}   \\ \midrule
Oracle extractive & \textbf{53.56}$_{\pm0.58}$           & \textbf{25.54} $_{\pm0.78}$         & \textbf{49.56}$_{\pm0.65}$   & 14.85 &  13.45  & 16.13      \\ 
BERT extractive              & 26.60$_{\pm0.51}$          & 11.11$_{\pm0.41}$          & 24.59$_{\pm0.47}$    & \textbf{13.44}  & \textbf{13.26}  &  \textbf{14.40}      \\ \midrule
Pointer generator                & 38.33$_{\pm0.61}$          & 14.11$_{\pm0.46}$          & 35.81$_{\pm0.60}$     &  16.36  &  15.86  &  15.90     \\ 
BART                 & 52.53$_{\pm0.51}$          & 21.83$_{\pm0.52}$          & 49.75$_{\pm0.52}$          &  13.59  &  14.16  &  14.45 \\ 
BART+CNN/DM             & 52.46$_{\pm0.48}$          & 21.84$_{\pm0.50}$          & 49.70$_{\pm0.50}$     & 13.73  &  14.33  &  14.60      \\ 
BART+PubMed            & 52.66$_{\pm0.48}$          & 21.73$_{\pm0.48}$          & 49.97$_{\pm0.51}$     &  \textbf{13.30}  &  \textbf{13.80}  &  \textbf{14.28}     \\ 
BART+CNN/DM+PubMed   & \textbf{53.02}$_{\pm0.48}$ & \textbf{22.06}$_{\pm0.49}$ & \textbf{50.24}$_{\pm0.49}$  &  13.60  &  14.11  &  14.41 \\ 
 \bottomrule
\end{tabular}
\caption{Test set performance evaluated by ROUGE and readability score. BART model pretrained on CNN/DM and PubMed is the best-performing model based on ROUGE, while BART model pretrained on PubMed is the best one based on readability score (Best model performance is in bold). $x_\pm$ indicates 95\% interval: $[x-, x+]$}
\label{rouge_readability}

\end{table*}

\subsection{Methods}
Summarization methods can be broadly categorized into extractive and abstractive approaches. The extractive approach 
creates summaries by selecting the most important sentences in a document,
while the abstractive approach usually employs sequence-to-sequence models to generate summaries that may contain new phrases not included in the source document. We experimented with several state-of-the-art extractive and abstractive methods to check the feasibility and difficulty of the plain language summarization task. 
\subsubsection{Extractive methods}
We applied two extractive methods -- \textit{Oracle extractive} and \textit{BERT extractive} \cite{liu2019text} -- to the CDSR dataset. \textit{Oracle extractive} can be viewed as an upper bound for extractive models. It creates an oracle summary by selecting the set of sentences in the document that generates the highest ROUGE-2 score with respect to the gold standard summary. Since oracle extractive summarization takes the gold standard summary into consideration, it can't be applied summarization tasks in practice. \textit{BERT extractive} is the state-of-the-art extractive method for text summarization. \textit{BERT} \cite{devlin2018bert} is a bidirectional unsupervised language representation derived by pre-training a Transformer architecture on a unlabeled text corpus for reconstruction. Several inter-sentence Transformer layers are then stacked on top of BERT outputs, to capture document-level features for extracting summaries. A sigmoid classifier is added as the output layer for extractive summarization. The oracle summary in the \textit{Oracle Extractive} method are used as supervision for training the \textit{BERT Extractive} model. 
\subsubsection{Abstractive methods}
We experimented with two abstractive models, \textit{pointer-generator} \cite{see2017get} and \textit{BART} \cite{lewis2019bart}, for our task. \textit{Pointer-generator} was a commonly used abstractive model before pretraining dominated the field. It enhances the standard sequence-to-sequence model with a pointer network that allows both copying words from the source and generating words from a fixed vocabulary. \textit{BART} is a state-of-the-art summarization model based on a large transformer sequence-to-sequence architecture. It is pre-trained on large corpora by corrupting text with an arbitrary noising function, and learning a model to reconstruct the original text.
As a sequence-to-sequence model, BART can be directly fine tuned for abstractive summarization task.
\subsubsection{Intermediate pre-training}
\label{sec:inter-pretraining}
To compensate for the limited training data, we added intermediate pre-training steps for the BART model before finetuning.
We first experimented with adding labeled data for summarization task in other domains. We adopted the CNN/DM dataset \cite{nallapati2016abstractive}, which contains about 287K document-summary pairs, and BART is among the best-performing systems for this task. 
Secondly, we tried to pre-train BART with an unlabeled biomedical corpus to expose the model to medical domain-specific language. We used the PMC articles dataset \footnote{https://www.kaggle.com/cvltmao/pmc-articles} 
which contains 300K PubMed abstracts. Following the BART paper, we corrupted these documents using several transformations, including text substitution and sentence shuffling. BART was then trained on the corrupted abstracts to reconstruct the original PubMed abstracts. Lastly, we combined these two strategies to train BART on CNN/DM and PubMed sequentially before finetuning it on our dataset.
\subsubsection{Training details}
All experiments were run using a single NVIDIA Tesla V-100 GPU. All models were developed using \texttt{PyTorch}. 
We used \texttt{neural-summ-cnndm-pytorch}\footnote{https://github.com/lipiji/neural-summ-cnndm-pytorch/} to implement the pointer-generator model. The batch size was set to 4. Other hyper-parameters were set to default values. We built the BERT extractive model using code released by the authors.\footnote{https://github.com/nlpyang/presumm} 
The learning rate was set to $2 \times 10^{-3}$ and the batch size 140. Other hyper-parameters were set to default values. We used the Fairseq \footnote{https://github.com/pytorch/fairseq} BART implementation. All BART models were trained using the Adam optimizer. The learning rate was set to $3 \times 10^{-5}$, and learning decay was applied. The minimum length of the generated summaries was set to 100, and the maximum length was set to 700.
\subsection{Results}
\subsubsection{Automated evaluation}


\begin{table*}
\centering
\begin{tabular}{@{} l cc cc @{}}
\toprule
                  & \multicolumn{2}{c}{\textbf{Abstract A}} & \multicolumn{2}{c}{\textbf{Abstract B}} \\
\cmidrule(lr){2-3}
\cmidrule(lr){4-5}
\textbf{Perspectives} & \textbf{Target}  & \textbf{Generated} & \textbf{Target}  & \textbf{Generated} \\
\midrule
Grammaticality            & 4.25    & 4.50        &3.50     & 4.00 \\
Meaning Preservation       & 3.75    & 4.75       & 3.50     & 4.50 \\
Understandability  & 3.75    & 3.50        & 2.75    & 2.50     \\
Correctness of Key Information        & 3.50     & 4.50    & 4.00       & 4.00 \\
\bottomrule
\end{tabular}
\caption{Human evaluation scores of the expert-generated summaries (\textit{Target}) and the model-generated summaries (\textit{Generated}) for two abstracts from the test set. Generated abstracts from BART+CNN/DM+PubMed model have better scores in grammaticality, meaning preservation, and correctness of key information.}
\label{result_human_evaluation}

\end{table*}
ROUGE and readability results on the CDSR test set are shown in Table \ref{rouge_readability}. We compare the seven methods described above: Oracle extractive, BERT extractive, pointer-generator, BART, BART pre-trained on CNN/DM, BART pre-trained on Pubmed abstracts, and BART pre-trained on both CNN/DM and PubMed abstracts. 

The oracle extractive method, as an upper bound for the extractive approach, produces the best ROUGE-1 and ROUGE-2 scores. However, it obtains approximately the same level of readability as the source text in our test set (Table \ref{data_description})
, which indicates that selecting the reference sentences will only result in a summary that is as difficult to read as the original abstract.
In contrast, the BERT-based extractive model achieves better readability scores while performing worst in terms of ROUGE scores. This demonstrates that, in practice, training the model to extract the correct content from the original abstract might be difficult, even though the model learns to extract shorter and easier sentences.

Among the 5 abstractive models, the pointer-generator model performs significantly worse in both ROUGE and readability, emphasizing the importance of pre-training for our task.
BART-based models achieve surprisingly good performance in terms of both summarization and readability, suggesting contemporary NLP models have the potential to perform the task, and to help the general public access professional medical information. Additionally, BART pre-trained on CNN/DM and PubMed abstracts achieves the best performance in ROUGE, and BART pre-trained only on PubMed abstracts obtained the lowest readability. 
This demonstrates the usefulness of either adding task-relevant labeled data or domain-specific unlabeled data. However, our strategies for adding such data are quite straightforward, and we lacked resources to do hyperparameter search for the relatively expensive pre-training procedure. Therefore, we only see marginal improvement compared with the BART model. We will aggregate more relevant data, and develop better pre-training strategies to improve the performance in future work.

\subsubsection{Human evaluation}
Table \ref{result_human_evaluation} shows the human evaluation results. Intriguingly, human evaluators rated the model-generated summaries with comparable or even higher scores for all the four aspects, and for both abstract A and B. The average Kendall’s coefficient \cite{sen1968estimates} for the two biomedical abstracts among all evaluators' inter-rater aggrement is 0.62. Kendall’s coefficient ranges from -1 to 1, indicating low to high association. Considering the subjectivity of the rating task, this number indicated high human agreement for the tasks. 
While larger scale study is required,
this work provides preliminary evidence that automatically-generated plain language summaries
are readable and interpretable to non-expert human readers.
\subsection{Qualitative analysis}
We present the output of our best two models in the last two columns of Table \ref{transformation_case_study}. 
This provides evidence that the best-performing models can address some transformations, and generate grammatical and meaningful outputs. 
Specifically, out of the five listed phenomena, we observed that model-generated summaries could achieve three transformation types to some extent, including removing unnecessary details, jargon explanation and sentence structure simplification. 
Some capabilities the model demonstrated are encouraging for future research. 
For example, it learned to explain the term RCT from similar examples in the training data.

On the downside, the  models are still struggle with some difficult transformations, such as relevant background explanation. This ability is harder to learn, and our dataset might not contain the required background knowledge. 
Therefore, external knowledge might be also useful. 
Furthermore, we also see risks in using the current abstractive models to generate reliable information for the public. For example, in the example of sentence structure simplification, \textit{BART+PubMed} changed the meaning of the original sentence: the source sentence claims an association between the pattern of blood flow with poor prognosis, while the generated sentence focuses on the Doppler ultrasonography. \textit{BART+CNN/DM+PubMed}
performs better in this case.


\section{Discussion}
Automated lay language summarization of biomedical scientific reviews requires both summarization and the acquisition of domain knowledge.
Previously, available datasets 
were constructed at sentence level. However, sentence-level simplification or transformation does not require the complex strategies used by experts when rendering biomedical literature understandable to a lay audience. Therefore, we consider the 
document-level dataset as an important outcome of our work,  which 
can be useful for future research on this topic.
Abstractive models are more practical than extractive ones, 
since extractive summaries 
are written
in the same professional language as their source documents. 
The best performing model is BART pre-trained on both CNN/DM and PubMed abstracts, which preserves key information (based on ROUGE) while dropping the reading requirements a year or two (based on readability scores).  

Human evaluation is necessary for our task.
There is a considerable gap between the automatic evaluation merics 
and human judgement. 
Despite being widely used to evaluate summarization systems, ROUGE is not practical for our task because it can neither capture the required transformation phenomena nor assess difficulty in understanding. 
Similarly, lower readability scores do not imply understandability. 
Readability scores consider only the surface forms,
without considering the complexity introduced by medical abbreviations and domain-specific concepts. 
Human evaluation is the most robust method to evaluate the performance. 
However, aside from the small number of participants, 
the survey questions need a formal validity.
Further studies are required to find that BART-derived summaries were more appealing to human raters on several fronts hold when more abstracts and human raters are involved.

\section{Conclusion}
We propose a novel plain language summarization task at the document level and construct a dataset to support training and evaluation. The dataset is of high quality, and the task is challenging due to typical transformation phenomena in this domain. We tried both extractive and abstractive summarization models, and obtained best performance with a BART model pre-trained further on CNN/DM and PubMed, as evaluated by automated metrics. Human evaluation suggests the automatically generated summaries may be at least as acceptable as their professionally authored counterparts.

\newpage
\bibstyle{aaai21}
\bibliography{main}
\end{document}